\pdfoutput=1

\documentclass[11pt]{article}
\usepackage{enumitem}
\usepackage{stfloats}
\usepackage[most]{tcolorbox}    
\tcbuselibrary{listings}  
\tcbuselibrary{breakable} 
\usepackage{listings}     
\usepackage{amsmath}      
\usepackage{amssymb}      
\usepackage{fontawesome5} 
\usepackage[final]{acl}
\usepackage{tabularx}
\usepackage{colortbl}   

\usepackage{makecell}
\usepackage{multirow}
\usepackage{xspace}
\newcommand{\paratitle}[1]{\vspace{1.5ex}\noindent\textbf{#1}}
\newcommand{\ie}{\emph{i.e.,}\xspace}

\newcommand{\eg}{\emph{e.g.,}\xspace}

\newcommand{\ignore}[1]{}
\usepackage{times}
\usepackage{latexsym}

\usepackage[T1]{fontenc}

\usepackage[utf8]{inputenc}

\usepackage{microtype}
\usepackage{CJKutf8}        

\usepackage{inconsolata}
\usepackage{booktabs}
\usepackage{graphicx}

\usepackage{algorithm}
\usepackage{algpseudocode}
\usepackage{amsmath}

\makeatletter
\def\@fnsymbol#1{}
\makeatother

\title{ManuSearch: Democratizing Deep Search in Large Language Models \\with a Transparent and Open Multi-Agent Framework}

\author{
  Lisheng Huang$^{1*}$\thanks{* Equal contribution.}~,
  Yichen Liu$^{1*}$,
  Jinhao Jiang$^{1*}$,\\
  \textbf{Rongxiang Zhang}$^{2}$,
  \textbf{Jiahao Yan}$^{1}$,
  \textbf{Junyi Li}$^{3\dagger}$ \thanks{$^{\dagger}$ Corresponding authors.}, 
  \textbf{Wayne Xin Zhao}$^{1\dagger}$\footnotemark[3] \\
  $^1$Gaoling School of Artificial Intelligence, Renmin University of China.\\
  $^2$Harbin Institute of Technology.\\
  $^3$Department of Computer Science, National University of Singapore.\\
  \texttt{\{huanglisheng, liuyichen\}@ruc.edu.cn} \\
  \texttt{batmanfly@gmail.com}
}

\begin{document}
\maketitle

\begin{abstract}

Recent advances in web-augmented large language models (LLMs) have exhibited strong performance in complex reasoning tasks, yet these capabilities are mostly locked in proprietary systems with opaque architectures. In this work, we propose \textbf{ManuSearch}, a transparent and modular multi-agent framework designed to democratize deep search for LLMs. ManuSearch decomposes the search and reasoning process into three collaborative agents: (1) a solution planning agent that iteratively formulates sub-queries, (2) an Internet search agent that retrieves relevant documents via real-time web search, and (3) a structured webpage reading agent that extracts key evidence from raw web content. To rigorously evaluate deep reasoning abilities, we introduce \textbf{ORION}, a challenging benchmark focused on open-web reasoning over long-tail entities, covering both English and Chinese. Experimental results show that ManuSearch substantially outperforms prior open-source baselines and even surpasses leading closed-source systems. Our work paves the way for reproducible, extensible research in open deep search systems. We release the data and code in \url{https://github.com/RUCAIBox/ManuSearch}

\end{abstract}

\section{Introduction}\label{sec:intro}

Deep search systems have recently marked remarkable strides by coupling large language models (LLMs) with web search, enabling them to answer complex queries that require multi-step reasoning and up-to-date information~\citep{open-deep-search}. Proprietary systems such as Perplexity's Sonar Reasoning Pro~\citep{perplexity-ai} and OpenAI's GPT-4o Search Preview~\citep{gpt-search-ai} exemplify this progress. These closed-source agents demonstrate emergent chain-of-thought reasoning capabilities: they autonomously plan out search queries, retrieve information from the web, and synthesize coherent, context-rich answers with source citations. 
By integrating search planning, iterative retrieval, and on-the-fly content aggregation, these agents achieve state-of-the-art performance on challenging benchmarks, far surpassing what static offline models can do.

Despite impressive capabilities of closed systems, the open-source ecosystem lacks comparable and transparent alternatives. Key components of deep search systems, \eg sophisticated query planning, multi-hop retrieval, and tool-augmented reasoning, remain entangled within proprietary stacks, with few modular open implementations. In particular, there is a lack of modular and interpretable architectures that let researchers inspect or improve each stage of the reasoning process and an absence of benchmarked multi-stage reasoning agents that can serve as open baselines. More importantly, there exists significant difficulty in replicating the performance of closed systems due to their restricted access and opaque design. This growing gap between private and public AI has been noted by the research community~\citep{open-deep-search,zheng2025deepresearcher}. Initial efforts to close it are only just emerging; for example, an early open-source prototype has combined techniques like ReAct-based tool use and prompting to nearly match the quality of GPT-4o Search and Sonar on certain tasks~\citep{open-deep-search}. However, these attempts are still in their infancy, underscoring the urgent need to democratize deep search research and foster reproducible innovation in this domain.

In this paper, we address these challenges by introducing \textbf{ManuSearch}, a transparent and open-source deep search system. Specially, ManuSearch is designed as an agent-based, modular system that achieves web-scale complex reasoning tasks with three collaborative agents: (1) \textit{Solution planning agent}, an LLM-based planner that interprets the user's query, formulates a strategy (a series of sub-questions or steps), and decides which information to seek at each step; (2) \textit{Internet search agent}, a specialized agent that takes the planner's requests, executes web searches, and gathers relevant evidence from the open Internet; and (3) \textit{Webpage reading agent}, an agent that reads the retrieved webpages, and extracts the most relevant key information needed to answer the query. These agents communicate and iterate in a structured reasoning loop, effectively integrating task planning, open Internet search, and key information comprehension and synthesis into a unified problem-solving process. By breaking the deep search pipeline into interpretable modules, ManuSearch provides an extensible and transparent alternative to monolithic closed-source systems. More important, each agent's behavior is traceable, \ie, one can examine the chain-of-thought in the planner's decisions, the queries issued, and the evidence used to support the answer, which not only aids debugging and trust, but also allows researchers to improve individual components in isolation.

To enable a rigorous evaluation of our system's deep search capabilities, we introduce \textbf{ORION}, a benchmark for \underline{O}pen-web \underline{R}easoning evaluat\underline{ION} specifically designed based on \emph{long-tail entities}. Unlike existing datasets that often concentrate on high-frequency topics, ORION emphasizes reasoning over less common entities across ten diverse domains. Each question in ORION is constructed using predefined reasoning templates, ensuring that the benchmark challenges a wide spectrum of cognitive abilities. Questions are initially generated using LLMs and then refined by human annotators to verify factual correctness and provide source-grounded reasoning chains. Spanning both English and Chinese samples, ORION contains 310 annotated examples, each linked to authoritative sources. Our evaluation shows that even leading closed-source systems achieve under 30\% accuracy on ORION, underscoring the benchmark’s difficulty and its value in advancing research on transparent and modular deep search systems.

We conduct extensive experiments on our benchmark and two challenging datasets (\ie FRAMES and GAIA) to verify the effectiveness of our approach. The experimental results show that our ManuSearch system significantly outperforms previous open-source deep search systems.

\section{Related Work}

\paratitle{Large Reasoning Models.} Large reasoning models (LRMs) such as OpenAI-o1~\cite{openai-o1} and DeepSeek-R1~\cite{deepseek-r1} demonstrate impressive long-horizon reasoning on complex tasks, but still rely on static internal knowledge, causing them to falter on knowledge-intensive queries. Recent efforts have applied reinforcement learning (RL) to further boost the reasoning prowess of these models~\cite{deepseek-r1,zeng2025simplerl}. RL fine-tuning has enabled LRMs to excel at decomposing complex problems into multi-step solutions, achieving strong performance in domains like mathematical proof and code generation~\cite{qin2024o1,abs-2412-00154}. Beyond improving internal reasoning, an emerging direction is to let LRMs plan and act in tandem. ReAct-style prompting exemplifies this integration of planning with execution: the model interleaves chain-of-thought reasoning steps with external tool use, dynamically querying resources mid-problem and incorporating new information into its reasoning. This synergy between reasoning and acting leads to more reliable, factual outcomes on challenging tasks, marking a significant advance in LRM capabilities. However, these methods are constrained by their reliance on static, parameterized architectures that lack access to external world knowledge.

\paratitle{Deep Search with LLMs.} To overcome the knowledge limitations of static models, a new class of deep web-integrated reasoning agents has emerged. Recent systems such as Search-o1~\cite{Search-o1}, WebThinker~\cite{Li2025WebThinker}, and Open Deep Search~\cite{open-deep-search} augment an LLM's reasoning by weaving in web search planning, tool use, and evidence retrieval as part of the pipeline. These agents decompose complex queries into search sub-tasks and iteratively gather information from the open web, feeding the retrieved evidence back into the model’s chain-of-thought. A key trend is the use of RL-based training to scale these deep research capabilities. DeepResearcher~\cite{zheng2025deepresearcher}, for instance, trains an LLM agent end-to-end in a live web environment via reinforcement learning, yielding emergent behaviors like plan formulation, cross-source verification, and self-correction during multi-hop research. Similarly, R1-Searcher~\cite{r1-searcher} uses a two-stage outcome-driven RL strategy to explicitly incentivize the model to invoke external search, significantly improving open-domain question answering even against strong retrieval-augmented baselines. This line of work demonstrates that by integrating search and reasoning in a coordinated pipeline, it is possible to substantially enhance the deep research abilities of an LLM.
\section{ORION}

Existing complex reasoning datasets focus primarily on constructing multi-domain questions to evaluate various capabilities of large language models \citep{krishna2024fact, rein2024gpqa}. Although these models perform well on questions involving common entities, their performance tends to degrade when handling questions that involve reasoning on long-tailed entities. In this section, we introduce {ORION}, a benchmark for the evaluation of open-web reasoning over long-tail entities.

\subsection{Data Source and Reasoning Patterns}

In the process of benchmark construction, we prioritize the selection of long-tail entities from multiple domains to ensure diversity, while also designing questions that require complex logical operations to enhance the benchmark's complexity.

\paratitle{Seed Entity Selection.}
To mimic real-world web scenarios, we systematically select entities from ten diverse common domains in web search, \ie music, sports, geography, art, politics, science, games, history, TV shows and business. Entity selection excludes ambiguously named subjects and requires each entity to possess at least three verifiable numerical or temporal attributes to support multifaceted reasoning.

\paratitle{Reasoning Pattern Design.}
To synthesize complex questions, we define five reasoning patterns through atomic operations that require multistep knowledge composition. Each pattern demands distinct cognitive capabilities while ensuring answer verifiability through deterministic computation or unambiguous factual lookup. Table~\ref{tab:reasoning_patterns} summarizes these patterns with illustrative examples.

\subsection{Example Synthesis and Annotation}
The construction of our benchmark integrates automatic question generation based on LLMs followed by human refinement. Initially, we adopt an LLM to synthesize questions based on predefined reasoning patterns and seed entities. After that, human annotators carefully verify each question by labeling multiple sources of evidence, ensuring that the entire reasoning chain leading to the correct answer is thoroughly recorded with authoritative references.

To ensure the difficulty of the final synthesized questions, we use advanced QA assistants (\eg ChatGPT) to answer those questions and retain those that challenge the assistants. For questions identified as relatively simple, we apply iterative revisions, including substituting high-frequency entities with long-tail alternatives from provided domains and enhancing the logical complexity of the questions by additional reasoning constraints.

\begin{figure}[t]
    \centering
    \includegraphics[width=0.85\linewidth]{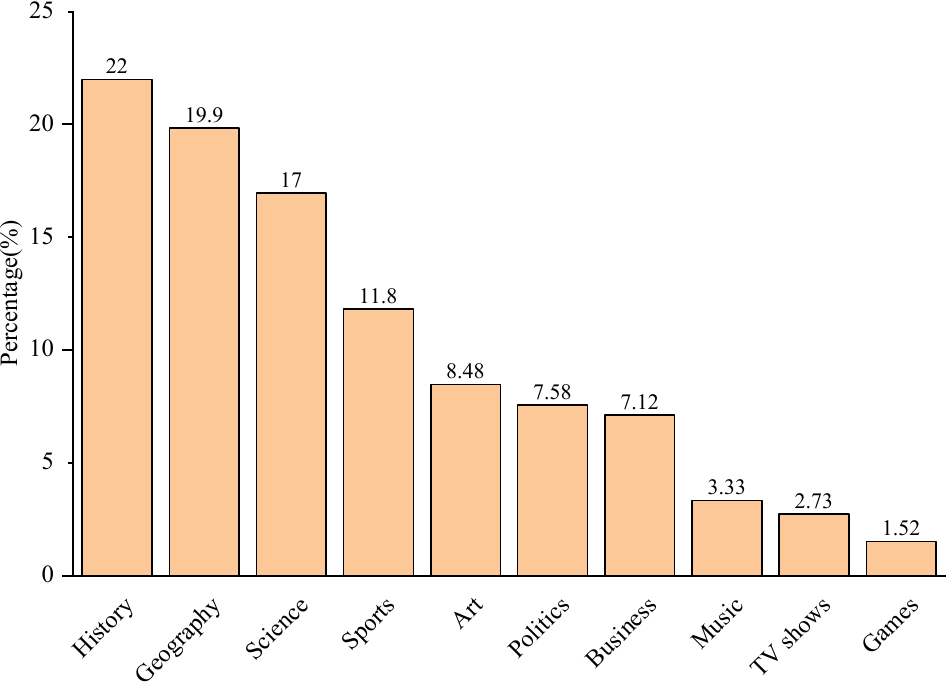}
    \caption{Distribution of domains in ORION.}
    \label{fig:distribution}
\end{figure}

\begin{table}[t]
\centering
\resizebox{\linewidth}{!}{
\begin{tabular}{lccc}
\toprule
       \textbf{AI Systems} & \textbf{Chinese} & \textbf{English} & \textbf{Overall} \\
\midrule
Kimi Exploration Edition & 14.7 & 20.0 &  17.1 \\
Doubao Search & 23.5 & 30.7 &  26.8 \\
Qwen2.5-Max Search & 20.0 & 20.7 & 20.3 \\
\bottomrule
\end{tabular}
}
\caption{Accuracy (\%) for three systems in ORION.}
\label{tab:mac-results}

\end{table}

\subsection{Benchmark Statistics}
Our benchmark comprises 310 samples, with 170 Chinese and 140 English entries. Each entry includes a question, a verifiable answer, and authoritative source URLs that illustrate evidence extraction.
The questions exhibit a diverse range of reasoning patterns, with more than 85\% involve more than two reasoning patterns, and more than 43\% involve more than three reasoning patterns. Moreover, they cover a broad spectrum of domains, as shown in Figure~\ref{fig:distribution}. The credibility of the sources is carefully verified, with more than 95\% URLs linking to authoritative data sources (\eg Wikipedia, academic articles, government documents). To substantiate the pronounced complexity of of our benchmark, we evaluate three state-of-the-art AI search systems on our benchmark, including Kimi Explorer, Doubao AI Search, and Qwen2.5-Max AI Search. Each system is tasked with answering questions by leveraging real-time search capabilities and information retrieval from the Internet. The results are presented in Table~\ref{tab:mac-results}. We can see that the three systems achieve accuracy rates below 30\%, which demonstrates the high difficulty of our dataset when dealing with questions involving intricate logical operations and rare entities.

\section{Method}\label{sec:method}

\subsection{Overview}

Previous work~\citep{r1-searcher,chen2025learning} typically relied on advanced capabilities of state-of-the-art reasoning models (\eg OpenAI-o1~\citep{openai-o1}, Deepseek-R1~\citep{deepseek-r1}), integrating both task decomposition and sub-problem solving within a single model. Such complex integration may exceed the capabilities of backbone models, often necessitating additional training or prompt engineering efforts. Moreover, performing planning and problem solving simultaneously requires the model to interact with the external web and integrate large volumes of web information, exceeding the model's context window.

To address these challenges, we decouple the traditional search paradigm and propose a transparent and open-source multi-agent deep search system, called \textbf{ManuSearch}. Our system is designed as an agent-based modular system that consists of three LLM-based collaborative agents: \textit{solution planning agent}, \textit{internet search agent} and \textit{webpage reading agent}. ManuSearch offers a plug-and-play deep search framework that supports flexible integration of any LLM, from open-source alternatives to commercial LLMs accessible via API. Next, we will describe each module in detail.

\begin{figure*}[t]  
\centering
\includegraphics[width=\textwidth]{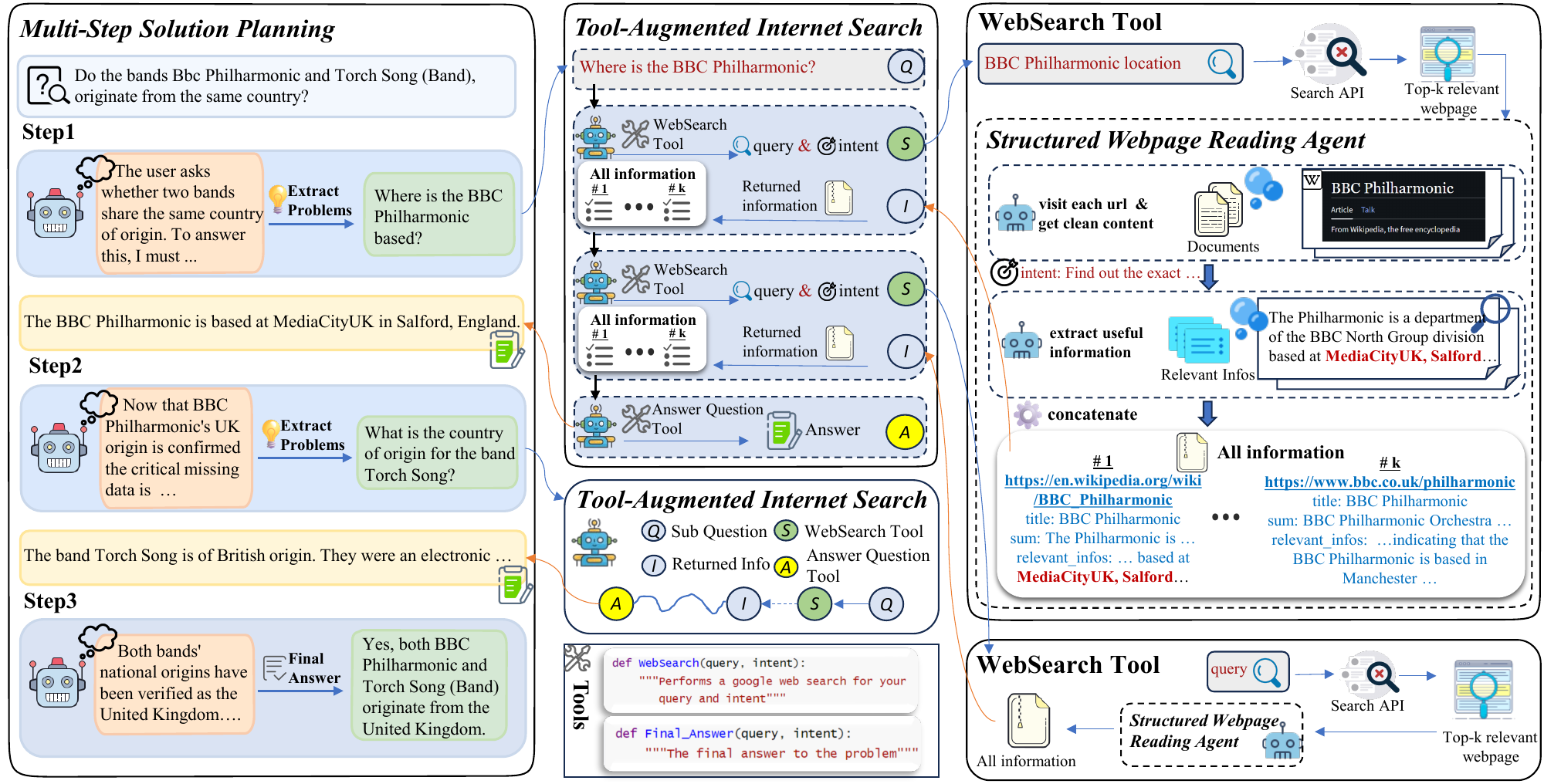}
\caption{The overall framework of our proposed ManuSearch.}
\label{fig:pdf-page}
\end{figure*}
 
\subsection{Multi-Step Solution Planning}\label{sec:3.2}

In complex problem-solving scenarios, particularly those that involve large search spaces and intricate dependencies, planning plays a pivotal role in guiding the reasoning process toward efficient and effective outcomes. In deep search, the system must determine not only what to search, but also how to structure the search process: identifying intermediate information needs, sequencing related subqueries, and reasoning over retrieved content.

To address this challenge, we design a specialized \textbf{Solution Planning Agent} that operates within a deep search framework. The planning agent solely focuses on breaking down the input problem iteratively and generating the final answer, granting the model superior capacity to reflect on and evaluate the main problem-solving process and the correctness of sub-question solutions. Specially, the planning agent adopts the ReAct architecture~\citep{yao2023react} and utilizes a memory container to manage its input context. 
The memory manager records each decomposed sub-question and its corresponding answer generated by the tool-augmented Internet search agent (detailed in Section~\ref{sec:3.3}), then automatically concatenates them into the context for the LLM in the next iteration. 

At step $t$, the input question $x$, previously decomposed sub-questions and their corresponding answers (if any) are combined as the context, denoted by $\mathcal{H}_{t-1}=\{x, \langle q_1,a_1\rangle\,\ldots, \langle q_{t-1},a_{t-1} \rangle\}$. The solution planning agent first evaluates the problem-solving progress in the history and then either decomposes the input question further or refines the unsolved sub-question from the previous steps, generating the next sub-question $q_{t}$ to be solved as: $q_{t} = \pi_{p}(\mathcal{H}_{t-1})$, where $\pi_{p}$ denotes the solution planning agent.
In particular, if the planning agent determines that the current information is sufficient to derive the final answer, the agent will proceed to generate the answer as: $y = \pi_{p}(\mathcal{H}_{t-1})$.

In ManuSearch, the planning agent only needs to generate the next-step query based on the environmental feedback (\ie previous sub-questions and solutions), without concerning itself with the details of the retrieval operations. This effectively alleviates the capability limitations faced in traditional retrieval-based frameworks.

\subsection{Tool-Augmented Internet Search}\label{sec:3.3}

After planning the next sub-question, we introduce a \textbf{Tool-Augmented Internet Search Agent}, responsible for solving sub-questions by invoking online Internet search. To ensure a unified manner with the planning agent, the Internet search agent also uses the ReAct format, interacting with the online Internet through pre-defined search tools in multiple rounds to ultimately solve sub-questions. We elaborate two tools to augment the search agent:  

$\bullet$ \textbf{Web Search:} 
The search tool will take a query and its corresponding search intent as input and performs information retrieval via Google API, returning the top-$K$ most relevant web pages information. Each result includes the link, title, short summary and detailed relevant information. 
Then, it will invoke the webpage reading agent (Section \ref{sec:3.4}) to visit the entire page documents of all retrieved links to extract detailed relevant information.

$\bullet$ \textbf{Answer Question:} This tool is called to generate an answer $a_t$ to the given sub-question $q_t$ based on the fine-grained and relevant information returned by the webpage reading agent.

Specifically, based on the sub-question $q_t$ and the historical information $\mathcal{H}_{t-1}$, the Internet search agent engages in a multi-round interaction with the vast web. At each round, the agent calls the web search tool to return top-$K$ results, denoted by $\mathcal{D} = \{\langle u_k,t_k,s_k,c_k \rangle \}_{k=1}^K$, where $u_k, t_k, s_k, c_k$ denote the page link, title, summary, and the task-relevant content, respectively. Note that since webpages contain massive and noisy information, the webpage reading agent also extracts the most relevant content $c_k$ for return. 
The search agent will iteratively search until the retrieved information is sufficient to answer the sub-question. All the retrieved results will be combined as a comprehensive and up-to-date source $\mathcal{O}_t = \{\mathcal{D}_{j}\}_{j=1}^{J}$, where $J$ is the number of iterations. Finally, the agent will call the answer question tool to generate an answer $a_t$ to the sub-question $q_t$ as follows:
\begin{equation}
    a_t = \pi_{s}(q_t, \mathcal{O}_t, \mathcal{H}_{t-1}),
\end{equation}
where $\pi_s$ denotes the search agent.
The generated answer $a_t$ will be passed to the solution planning agent, continuing the next round of planning and problem solving. Through this iterative process, the two agents collaborate interactively until the input question is successfully solved. We argue that this decoupled planning-solving framework is particularly well-suited for addressing web-based complex reasoning tasks, as it can fully leverage the respective strengths of each module to achieve the optimal performance.

\subsection{Structured Webpage Reading}\label{sec:3.4}

On the Internet, there are a huge number of HTML webpages with messy organizational forms and encoding formats, making it difficult to process in a unified approach and seriously affects the model's comprehension on the webpage content. Besides, the webpages contain substantial redundant and noisy information that is useless for problem solving. Therefore, we design a \textbf{Structured Webpage Reading Agent} which is primarily responsible for two tasks: extracting clean texts from the messy HTML page and further extracting the most relevant information from noisy raw texts.

\paratitle{Messy HTML Formats Processing.} 

For each input query, the web search tool returns the top-$K$ webpages with their links, titles, and summaries, denoted by $\mathcal{D} = \{\langle u_k,t_k,s_k\rangle \}_{k=1}^K$. For each page link, the webpage reading agent first crawls the whole HTML page and then removes the HTML tags, special characters to obtain clean texts $z_k$.

\paratitle{Relevant Information Extraction.} 
The raw webpage texts still contain massive noisy information. Therefore, we enable the Internet search agent to generate a detailed search intent $I$ to help the webpage reading agent extract the most relevant content. Unlike the simple query fed into the search engine, the search intent incorporates broader contextual information, aiming to bridge the semantic gap between query and search results. Given the search intent $I$ and the processed clean raw text $z_k$, the webpage reading agent extracts highly relevant information as:  
\begin{equation}
    c_k = \pi_{r}(z_k, I),
\end{equation}
where $\pi_r$ denotes the webpage reading agent. Finally, the reading agent combines the extracted information with the webpage and returns to the Internet search agent.
The search agent might iterate the search process for several times by interacting with the webpage reading agent. In this process, the webpage reading agent effectively filters out irrelevant information while accurately extracting task-relevant details. This significantly mitigates information conflicts across webpages in Internet search agent's reasoning, enabling more efficient and effective information retrieval.

\section{Experiments}
\label{sec:experiments}
\subsection{Experimental Settings}
\label{sec:settings}

\paratitle{Datasets and Metrics.}
For evaluation, we select two complex multi-hop reasoning benchmarks, \ie FRAMES~\citep{krishna2024fact} and GAIA~\citep{rein2024gpqa}, along with ORION to evaluate model performance. Specially, FRAMES is intended to test the information retrieval and factual reasoning capabilities from single-hop to multi-hop questions, and GAIA focuses on challenging information retrieval tasks in general scenarios. We select the whole dataset of FRAMES and adopt the same evaluation subset of GAIA used in WebThinker~\citep{Li2025WebThinker}. For the three datasets, we adopt LLM-as-Judge with ChatGPT-4o as the evaluator to compare model outputs with ground truth and report Pass@1 accuracy.

\paratitle{Baselines.} 
We compare \textbf{ManuSearch} to the following three types of baselines:

$\bullet$ \textbf{Closed-source Search Systems}
include Perplexity Sonar Reasoning Pro~\citep{perplexity-ai} from Perplexity, GPT-4o Search Preview~\citep{gpt-search-ai} from OpenAI, Kimi Search\footnote{https://www.kimi.com/}, and Exa Search Pro\footnote{https://exa.ai/}. These are state-of-the-art systems with access to search engines.

$\bullet$ \textbf{Open-source Search Systems} include Open Deep Search (ODS)~\citep{open-deep-search}, Search-o1~\citep{Search-o1}, WebThinker~\citep{Li2025WebThinker}, and SimpleDeepSearcher~\citep{zheng2025deepresearcher}. 

$\bullet$ \textbf{Direct Reasoning} employs GPT-4o, Qwen2.5-Instruct-32B, QwQ-32B and DeepSeek-R1 to directly reason and generate answers without access to search engines. 

\begin{table*}[tbh]
    \centering
    \resizebox{0.95\linewidth}{!}{\begin{tabular}{lcccccccc}
    \toprule
    \multirow{2}{*}{\textbf{Method}} & \multicolumn{1}{c}{\textbf{FRAMES}} & \multicolumn{3}{c}{\textbf{ORION}} &  \multicolumn{4}{c}{\textbf{GAIA}} \\
    \cmidrule(lr){3-5} \cmidrule(lr){6-9} 
    & Avg. & EN & ZH & Avg. & Level 1 & Level 2 & Level 3 & Avg. \\
    \midrule
    \rowcolor{gray!20}
    \multicolumn{9}{l}{\textbf{Direct Reasoning (w/o Retrieval)}} \\
    Qwen2.5-32B-Instruct &24.5 & 14.3 & 11.2 & 12.8 & ~~20.5* & ~~9.6* & ~~8.3* & ~~13.6* \\
    QwQ-32B &32.5 & 27.1 & 23.5 & 25.3 & ~~30.8* & ~~15.4*  & ~~25.0*  & ~~22.3*  \\
    DeepSeek-R1& ~~30.1*  &37.9 & 35.3 & 36.6  & ~~43.6* & ~~26.9* & ~~8.3* & ~~31.1*  \\
    GPT-4o & ~~50.5* & 27.1 & 20.0 & 23.6 & 25.6 & 15.4 & 0 & 17.5  \\
    \midrule
    \rowcolor{blue!10}
    \multicolumn{9}{l}{\textbf{Closed-source Search Systems}} \\
    GPT-4o Search Preview & ~~65.6* & 38.6 & 22.9 & 30.8 & 48.7  & 17.3 & 0 & 27.1 \\
    Perplexity Sonar Reasoning Pro & ~~44.4* & 26.4 & 19.4 & 22.9 & 43.6 & 25 & 16.7 & 31.1  \\
    Kimi Search & 35.9 & 14.3 & 11.2  & 12.8 & 35.9 & 21.2 & 8.3 & 24.3  \\
    Exa Search Pro & 32.8 & 8.6 & 11.8 & 10.7 & 30.8 & 9.6  & 0 & 16.5  \\
    \midrule
    \rowcolor{yellow!20}
    \multicolumn{9}{l}{\textbf{Open-source Search Systems}}\\
        ODS-v1+DeepSeek-R1  & ~~56.7* & -  & - & - & - & - & - & - \\
    ODS-v2+DeepSeek-R1  & \textbf{~~75.3*} & - & - & - & - & - & - & - \\
    Search-o1-32B & ~~64.4* & - & - & -  & ~~53.8*  & ~34.6*  & ~~16.7*  & ~~39.8*  \\
    WebThinker-32B-Base & - & - & - & - & ~~53.8*  & ~~44.2* & ~~16.7* & ~~44.7* \\
    SimpleDeepSearcher & ~~68.7* & - & -  & - & ~~61.5* & ~~44.2* & ~~16.7* & ~~47.6* \\
    ManuSearch-QwQ-QwQ (Ours)  & 68.4 & 40.0 & 34.7 & 37.4 & 59.0 & 42.3 & \textbf{25.0} & 46.6 \\
    ManuSearch-R1-V3 (Ours) & 71.8 & \textbf{47.9} &\textbf{37.1}& \textbf{42.5} & \textbf{64.1}&\textbf{44.2} & 8.3 & \textbf{47.6} \\
    \bottomrule
    \end{tabular}
    }
    \caption{Main results on three challenging reasoning tasks: FRAMES, ORION, and GAIA, evaluating models on both closed-source and open-source search systems. The results are measured using the Pass@1 metric, with breakdowns for individual languages (EN, ZH) and hierarchical reasoning levels in GAIA. \textbf{Bold} fonts indicate the best performance among open-source models, and asterisks (*) denote results collected from other studies.}
    \label{tab:main_results}
\end{table*}

\paratitle{Implementation Details.}
We implement ManuSearch upon Qwen and DeepSeek series models. In Qwen series, we adopt QwQ-32B as the multi-step solution planning agent and Internet search agent, represented as \textbf{ManuSearch-QwQ-QwQ}. In DeepSeek series, we adopt DeepSeek-R1 and DeepSeek-V3 as the planning and search agent, respectively, represented as \textbf{ManuSearch-R1-V3}. In both series, we uniformly employ Qwen2.5-32B-Instruct as the base model for the webpage reading agent. We also conduct experiments with respect to the base model seleciton of webpage reading agent in Section~\ref{sec:5.3}. For Internet search agent, we set the maximum number of sub-queries per search to 3 when calling web search tool. For webpage reading agent, we set the maximum length of the original web text to 64K. For the web search tool, we retrieve top-$5$ relevant webpages. Specifically, when the ManuSearch fails to provide an answer, we switch to the direct resoning mode to generate the final answer.

\subsection{Main Results}
Table~\ref{tab:main_results} shows the result of ManuSearch and other baselines across three representative benchmarks. 

Firstly, it can be observed that ManuSearch is a general search framework for seamlessly augmenting any LLMs from 32B models to large closed-source reasoning models. For example, DeepSeek-R1 achieves 30.1\% accuracy on FRAMES, 31.1\%  accuracy on GAIA and 36.6\% accuracy on ORION. After pluging into ManuSearch, these performances surge to 71.8\% on FRAMES, 47.6\% on GAIA and 42.5 \% on ORION, achieving impressive performance gains and validating the efficiency of our designed search framework.

Secondly, with a modular, multi-agent framework, ManuSearch nearly matches existing state-of-the-art baselines on the two benchmarks: FRAMES and GAIA. For 32B models, ManuSearch-QwQ-QwQ achieves 46.6\% on GAIA, surpassing Webthinker by 1.9\% and 68.4\% on FRAMES, almost matching the elaborately trained search agent SimpleDeepSearcher. Moreover, ManuSearch-R1-V3 achieves advanced performance, with 47.6\% on GAIA and 71.8\% on FRAMES, nearly matching ODS-v2+DeepSeek-R1.

Finally, ManuSearch outperforms all closed-source Search AIs in three datasets. Notably, with only 32B models, ManuSearch improves the best existing baseline of the GPT-4o Search Preview by 2.8\% in accuracy on FRAMES, 6.6 \% on ORION and 19.5\% on GAIA. 

\subsection{Further Analysis}
\label{sec:5.3}

We report further analysis on FRAMES with randomly selected 100 samples and GAIA, due to the constraint of computational resources.

\paratitle{How to harmonize reasoning and non-reasoning models?}
Our system comprises three LLM-based collaborative agents where each agent supports various configurations by integrating different LLMs, enabling diverse combinations tailored to specific tasks. 
In this part, we aim to explore how reasoning and non-reasoning models can be harmonized in our system. 
Inspired by the human problem-solving process where people typically engage in more deliberate thinking during the problem decomposition phase and adopt faster, more intuitive thinking during the information search phase, we compare two configurations: (1) reasoning models for both solution planning and internet search agents; and (2) reasoning models for solution planning agent while non-reasoning models for internet search agent. We evaluate two reasoning models (\ie QwQ and DeepSeek-R1) and non-reasoning models (\ie Qwen2.5-32B-Instruct and DeepSeek-V3). The results are shown in Table~\ref{tab:analysis1}. We can observe that the performance differences between these configurations are relatively small. Within the Qwen series, using QwQ as both agents yields better results, whereas in the DeepSeek series, using DeepSeek-R1 paired with DeepSeek-V3 performs better. Our in-depth analysis reveals that due to the relatively weaker capabilities of the Qwen series models, the non-reasoning model sometimes struggles to identify key information from the content returned by the webpage reading agent during subproblem solving, resulting in poorer performance. In contrast, for the DeepSeek series, both reasoning and non-reasoning models are sufficiently capable of handling subproblem solving effectively. In this case, the performance bottleneck is primarily determined by the quality of the information returned by the webpage reading agent.

\begin{table}[ht]
    \centering
    \resizebox{\linewidth}{!}{\begin{tabular}{l c c c c c}
    \toprule
    \multirow{2.5}{*}{\textbf{Method}} & \multicolumn{1}{c}{\textbf{FRAMES}} & \multicolumn{4}{c}{\textbf{GAIA}} \\
    \cmidrule(lr){3-6} 
     & Avg. & Level 1 & Level 2 & Level 3 & Avg\\
    \midrule
    MS-QwQ-Qwen & 63.0 & 51.3 & 44.2 & 8.3 & 42.7    \\
    MS-QwQ-QwQ  & 64.0 & 59.0 & 42.3 & 25.0 & 46.6   \\
    MS-R1-V3 & 62.0 & 64.1 & 44.2 & 8.3 & 47.6 \\
    MS-R1-R1 & 63.0 & 64.1 & 44.2 & 0 & 46.6 \\
    \bottomrule
    \end{tabular}
    }
    \caption{Performance comparison of ManuSearch with four different configurations on FRAMES and GAIA.}
    \label{tab:analysis1}
\end{table}

\paratitle{Does webpage reading agent really need page selection?}
Through extensive experiments, we found that the implementation of the webpage reading pipeline has a significant impact on the overall performance of our framework. Inspired by previous work~\cite{chen2024mindsearch,open-deep-search,Li2025WebThinker}, we summarize two mainstream webpage reading approaches:

\begin{itemize}[leftmargin=2em]

    \item \textbf{Selective Reading:} The model autonomously selects appropriate webpages to read in detail as it needs.

    \item \textbf{Full Reading:}  The model does not perform page selection and instead uses the full content of all webpages retrieved by search engine.

\end{itemize}

In the first approach, the model is minimally affected by irrelevant web content and can focus on pages it deems potentially useful, allowing for a more deliberate reasoning process. However, it is often difficult for the model to accurately choose the right pages to read based solely on metadata, leading to information loss and increasing errors.
The second approach is more straightforward by allowing the model to read all potentially relevant content, and it significantly reduces the chance of missing critical information. Nevertheless, the model may struggle to identify the correct information amid a large volume of content, increasing the risk of hallucination. Table~\ref{tab:webpage_pipelines} presents the results of our experiments over the two methods. It can be observed that Full Reading achieves better overall performance, with significantly better results than Selective Reading under the ManuSearch with QwQ-QwQ configuration. Moreover, considering that the Full Reading approach is simpler and more efficient, we ultimately recommend and adopt the Full Reading method in our framework.

\begin{table}[ht]
    \centering
    \resizebox{\linewidth}{!}{\begin{tabular}{l c c c c c}
    \toprule
    \multirow{2.5}{*}{\textbf{Method}} & \multicolumn{1}{c}{\textbf{FRAMES}} & \multicolumn{4}{c}{\textbf{GAIA}} \\
    \cmidrule(lr){3-6} 
    & Avg. & Level 1 & Level 2 & Level 3 & Avg\\
    \midrule
    \textbf{MS-QwQ-Qwen}  \\
    Full Reading & 63.0 & 51.3 & 44.2 & 8.3 & 42.7     \\
    Selective Reading  & 62.0 & 61.5 & 42.3 & 0 & 44.7 \\
     \midrule
    \textbf{MS-QwQ-QwQ}  \\
    Full Reading   & 64.0 & 59.0 & 42.3 & 25.0 & 46.6  \\
    Selective Reading & 56.0 & 48.7 & 40.4 & 16.7 & 40.8 \\
    \bottomrule
    \end{tabular}
    }
    \caption{Performance comparison of Full Reading and Selective Reading on FRAMES and GAIA.}
    \label{tab:webpage_pipelines}
\end{table}

\paratitle{What are the impacts of different webpage reading models?}
Since the retrieved webpages often contain various types of noise, lengthy content, or present conflicting information, it poses significant challenges for models to extract useful information. Meanwhile, as discussed in the previous part, it is crucial to investigate how different models as webpage reading agents affect the performance of overall framework. We conduct experiments using three models as webpage reading agent, including Qwen2.5-32B-Instruct, QwQ-32B, and ChatGPT-4o-mini, under the ManuSearch with QwQ-Qwen and QwQ-QwQ configurations. The results in Table~\ref{tab:impact_of_summarization} shows that Qwen2.5-32B-Instruct, as a webpage reading model, achieves consistently leading performance on both dataset, except under the ManuSearch-QwQ-Qwen configuration on GAIA, where ChatGPT-4o-mini has a better performance. These results demonstrate that Qwen2.5-32B-Instruct is the most suitable model among the three for serving as a webpage reading agent.

\begin{table}[ht]
    \centering
    \resizebox{\linewidth}{!}{\begin{tabular}{l c c c c c}
    \toprule
    \multirow{2.5}{*}{\textbf{Method}} & \multicolumn{1}{c}{\textbf{FRAMES}} & \multicolumn{4}{c}{\textbf{GAIA}} \\
    \cmidrule(lr){3-6} 
    & Avg. & Level 1 & Level 2 & Level 3 & Avg\\
    \midrule
    \textbf{MS-QwQ-Qwen with}  \\
     Qwen2.5-32B-Instruct & 63.0 & 51.3 & 44.2 & 8.3 & 42.7     \\
     QwQ-32B & 59.0 & 59.0 & 25.0 & 8.3 & 35.9 \\
     ChatGPT-4o-mini & 55.0 & 46.2 & 48.0 & 16.7 & 43.7  \\
    \midrule
    \textbf{MS-QwQ-QwQ with}  \\
     Qwen2.5-32B-Instruct & 64.0 & 59.0 & 42.3 & 25.0 & 46.6    \\
     QwQ-32B  & 61.0 & 56.4 & 44.2 & 16.7 & 44.7  \\
     ChatGPT-4o-mini  & 59.0 & 48.7 & 32.7 & 25.0 & 37.9  \\
    \bottomrule
    \end{tabular}
    }
    \caption{Average accuracy comparison of ManuSearch with three kinds of webpage reading models: Qwen2.5-32B-Instruct, QwQ-32B, and ChatGPT-4o-mini, evaluated on FRAMES and GAIA benchmarks.}
    \label{tab:impact_of_summarization}
\end{table}

\section{Conclusion}

We present ManuSearch, a transparent and modular multi-agent framework that enables large language models to perform deep web-integrated reasoning. By decoupling the problem-solving process into three specialized agents, \ie solution planning, internet search, and structured webpage reading, our system promotes interpretability, extensibility, and performance. Extensive evaluations on our proposed benchmark ORION demonstrate that ManuSearch significantly outperforms prior open-source systems and rivals or exceeds several closed-source commercial agents. Our work highlights the importance of modular reasoning pipelines and introduces a reproducible foundation for future research in open deep search systems. We hope this framework will catalyze progress toward trustworthy agents empowered with search capabilities.

\section{Limitations}
Despite our considerable efforts, this study remains limited due to the substantial computational cost. Our evaluations primarily focus on open-source models from the Qwen and DeepSeek series and are conducted on only three datasets. Future research will expand the scope by incorporating a wider range of both open-source and proprietary models—such as the LLaMA 4 series~\citep{llama-4}, the GPT series~\citep{gpt-4.1}, and the Claude series~\citep{claude}—as well as evaluating performance across a more diverse set of datasets. Additionally, our current framework employs a fine-grained WebSearch tool as part of the Tool-Augmented Internet Search mechanism. Future work could enhance this framework by integrating additional built-in tools such as Code Execution and Multimodal tools to endow models with more comprehensive and versatile capabilities.

\section{Ethics and Risks}

Our work primarily focuses on the construction of the open-source deep search system ManuSearch and the open-web reasoning evaluation dataset ORION. Although our goal is to pave the way for reproducible and extensible research in open deep search systems, we recognize that it could be misused in certain scenarios, such as large-scale web scraping without proper authorization, automated generation or manipulation of online content, or privacy-invading search behaviors. To mitigate these risks, we clearly state the intended use of the framework, and we do not include any functionality for bypassing access restrictions or automated user impersonation. We encourage the responsible use of this framework and its continued critical evaluation.

\bibliography{custom}
\clearpage
\newpage
\appendix

\section{Appendix}
\label{sec:appendix}

\subsection{Case Study}

\begin{figure*}[!t]  
\centering
\includegraphics[page=1, trim=1cm 4cm 1cm 1.5cm, clip,width=\textwidth]{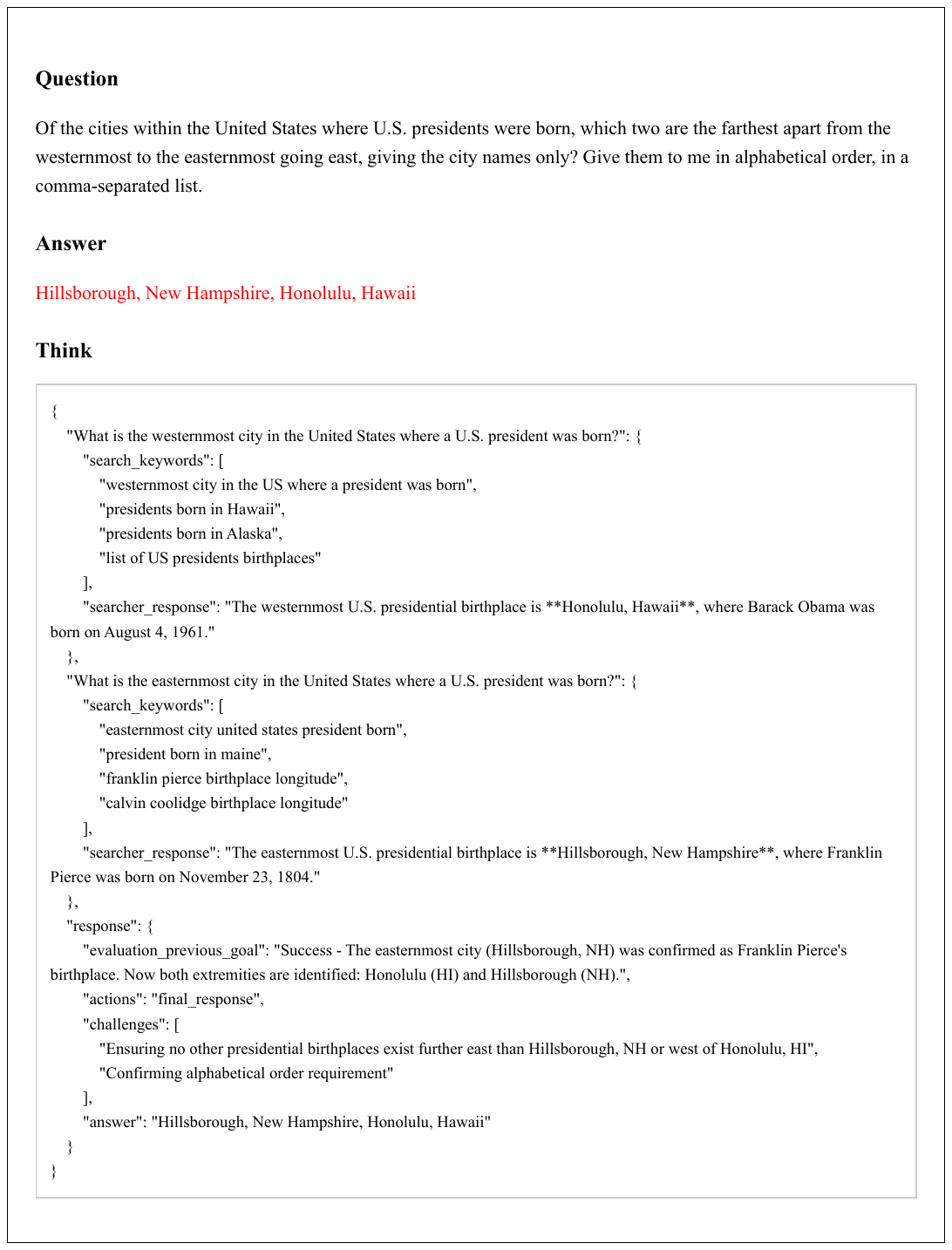}
\caption{A case of ManuSearch from GAIA benchmark.}
\label{fig:pdf-page}
\end{figure*}

In the case shown in Figure \ref{fig:pdf-page}, our search strategy begins by identifying the birthplaces of U.S. presidents and their geographical coordinates. At the first step, the \textbf{Solution Planning Agent} generates the current sub-question, querying "the westernmost city in the United States where a U.S. president was born". Upon receiving the sub-question, the \textbf{Tool-Augmented Internet Search Agent} produces a diverse set of search keywords, including "westernmost city in the US where a president was born", "presidents born in Hawaii", "presidents born in Alaska", and "list of US presidents birthplaces", to obtain comprehensive information. Subsequently, the Tool-Augmented Internet Search Agent calls the \textbf{WebSearch Tool} to retrieve the top-$K$ relevant documents, which are then passed to the \textbf{Structured Webpage Reading Agent} for processing. The Structured Webpage Reading Agent cleans and formats these documents, extracting content relevant to the query and returning it to the Tool-Augmented Internet Search Agent. Based on the extracted information and the user's query, the Tool-Augmented Internet Search Agent formulates an answer to the current sub-question. This iterative process continues until the Solution Planning Agent accumulates sufficient information to respond to the user's original question.

\begin{table*}[t]
\small
\centering
\begin{tabularx}{0.89\textwidth}{cX}
\toprule
\textbf{Reasoning Type} & \textbf{Description and Example} \\
\midrule
\textbf{Numerical Computation}  &  Combines arithmetic operations with numerical facts. \\
& \textit{Example: "Among TIME's Persons of the Year from 2000 to 2010, which tech industry winner born in an odd year had the square root of their age at election closest to an integer?"} \\
\midrule
\textbf{Temporal Constraint}  & Resolves time-bound relationships through duration calculations or chronological ordering. \\
& \textit{Example: "When Robert D. Heaton was born, how many years had Pennsylvania been part of the U.S.?"} \\
\midrule
\textbf{Fact Constraint} & Identifies entities satisfying more than two fact constraints through Boolean conjunction. \\
& \textit{Example: "Which U.S. president signed climate-related legislation while in office and was born 40–50 years before Halley's Comet's last return?"} \\
\midrule
\textbf{Statistical Reasoning} & Applies comparative operators or aggregation over bounded entity sets. \\
& \textit{Example: "How many UK heirs died with age more than 80 between 1707 and 2025?"} \\
\midrule
\textbf{Scenario Reasoning} & Embeds real entities into hypothetical scenarios, requiring real-world information for answering. \\
& \textit{Example: "Assuming the height of "Aurora Lab" (in meters) equals the first three digits of the speed of light (m/s) plus 200, what is its ranking among the world's tallest buildings in 2024?"} \\
\bottomrule 
\end{tabularx}
\caption{Reasoning patterns and illustrative examples in our benchmark ORION.}
\label{tab:reasoning_patterns}
\end{table*}

\subsection{Reasoning Patterns in ORION}
To synthesize complex questions, we define five reasoning patterns through atomic operations that require multi-step knowledge composition. Table~\ref{tab:reasoning_patterns} summarizes these patterns with illustrative examples.

\end{document}